\documentclass[twocolumn,numbook,natbib]{svjour3}  %

\usepackage[normalem]{ulem}
\usepackage{url}
\usepackage[breaklinks=true,colorlinks,citecolor=royalblue,bookmarks=false]{hyperref}
\usepackage{amsmath,amssymb,amsfonts}
\usepackage{bbm}
\usepackage{cleveref} 
\usepackage{graphicx}
\usepackage{textcomp}
\usepackage{xcolor}
\usepackage{stfloats}
\usepackage{multirow}
\usepackage{paralist}
\usepackage{tabularx}
\usepackage{tcolorbox}
\usepackage{booktabs}
\usepackage{subcaption}
\usepackage{natbib}

\usepackage{enumitem}

\definecolor{dg}{rgb}{0,0.694,0.298}
\definecolor{purple}{rgb}{0.4,0.176,0.569}
\definecolor{royalblue}{RGB}{65,105,225}
\definecolor{googleblue}{RGB}{66,133,244}
\definecolor{googlered}{RGB}{234,67,53}
\definecolor{googleyellow}{RGB}{251,188,5}
\definecolor{googlegreen}{RGB}{52,168,83}

\usepackage[normalem]{ulem}

\newcommand{\delete}[1]{}

\newcommand{\circled}[1]{\textcircled{\raisebox{-.9pt}{#1}}}
\usepackage{xspace}
\makeatletter
\DeclareRobustCommand\onedot{\futurelet\@let@token\@onedot}
\def\@onedot{\ifx\@let@token.\else.\null\fi\xspace}

\makeatother

\journalname{}
\makeatletter
\def\makeheadbox{}
\makeatother

\begin{document}

\title{Can Distillation Mitigate Backdoor Attacks in Pre-trained Encoders?}

\author{$\text{Tingxu Han}^{1,\dagger}$ \and
        $\text{Wei Song}^{2,\dagger}$ \and
        $\text{Weisong Sun}^{3}$ \and
        $\text{Ziqi Ding}^{2}$ \and
        $\text{Yebo Feng}^{3}$ \and
        $\text{Chunrong Fang}^{1,*}$ \and
        $\text{Jun Li}^{4}$ \and
        $\text{Hanwei Qian}^{1}$ \and Zhenyu Chen$^{1}$ \and Yang Liu$^{3}$      
}

\institute{ 
            \textsuperscript{\rm 1}Tingxu Han, Chunrong Fang, Hanwei Qian, and Zhenyu Chen are with the State Key Laboratory for Novel Software Technology, Nanjing University, Nanjing 210093, China (e-mail: \{txhan, qianhanwei\}@smail.nju.edu.cn, \{fangchunrong, zychen\}@nju.edu.cn). Zhenyu Chen is also with Shenzhen Research Institute of Nanjing University, China. \\
            \textsuperscript{\rm 2}Ziqi Ding and Wei Song are with the School of Computer Science and Engineering, University of New South Wales, New South Wales 2052, Australia (e-mail: ziqi.ding1@unsw.edu.au, wei.song1@unsw.edu.au). \\
            \textsuperscript{\rm 3} Weisong Sun, Yebo Feng, and Yang Liu are with the School of Computer Science and Engineering, Nanyang Technological University, Nanyang 639798, Singapore (e-mail: \{weisong.sun, yebo.feng, yangliu\}@ntu.edu.sg). \\
            \textsuperscript{\rm 4} Jun Li is with the Department of Computer Science and the Director of the Network \& Security Research Laboratory at the University of Oregon, America (e-mail: lijun@cs.uoregon.edu).  \\
            \textsuperscript{\rm $\dagger$} Both authors contributed equally to this work.  \\
            \textsuperscript{\rm *} Corresponding author: Chunrong Fang (Email: fangchunrong@nju.edu.cn). \\
}

\date{}

\maketitle

\begin{abstract}
Self-Supervised Learning (SSL) has become a prominent paradigm for pre-training encoders to learning general-purpose representations from unlabeled data and releasing them on third-party platforms for broad downstream deep learning tasks. 
However, SSL is vulnerable to backdoor attacks, where an adversary may train and distribute poisoned pre-training encoders to contaminate the downstream models.

In this paper, we study a defense mechanism based on distillation against poisoned encoders in SSL. Traditionally, distillation transfers knowledge from a pre-trained teacher model to a student model, enabling the student to replicate or refine the teacher's learned representations. We repurpose distillation to extract benign knowledge and remove backdoors from a poisoned pre-trained encoder to produce a clean and reliable pre-trained model. We conduct extensive experiments to evaluate the effectiveness of distillation in mitigating backdoor attacks on pre-trained encoders. Based on two state-of-the-art backdoor attacks and four widely adopted image classification datasets, our results demonstrate that distillation reduces the attack success rate from 80.87\% to 27.51\%, with only a 6.35\% drop in model accuracy. Furthermore, by comparing four teacher architectures, three student models, and six loss functions, we find that the distillation with fine-tuned teacher networks, warm-up-based student training, and attention-based distillation losses yield the best performance.

\keywords{Self-supervised Learning, Pre-trained Encoder, Backdoor Attack and Defense}

\end{abstract}

\section{Introduction}
\label{sec:intro}

With the rapid advancement of deep learning, self-supervised learning (SSL) has become increasingly popular~\citep{chen2020simple,he2022masked,caron2021emerging,lan2019albert,khosla2020supervised,baevski2020wav2vec}.
Following a ``pre-train first, fine-tune later'' paradigm, SSL pre-trains an encoder on large-scale, uncurated, and unlabeled data to learn general-purpose representations. The pre-trained encoder is then released on third-party platforms such as HuggingFace and ModelZoo~\citep{balestriero2023cookbook,wang2022ssl4eo}, where it can be downloaded and fine-tuned with task-specific data to support various downstream applications.
This paradigm greatly reduces the cost and effort required to train a new model for every individual task~\citep{jaiswal2020survey,ning-etal-2025-mode,xin-etal-2024-beyond}.

However, SSL is vulnerable to backdoor attacks \citep{2022-BadEncoder, BASSL,han2025mutual}. This vulnerability stems from the SSL paradigm, which relies on large-scale, uncurated datasets and widely shared pre-trained encoders—both of which expand the attack surface, allowing adversaries to compromise released encoders or tamper with pre-training data.
At the encoder level, adversaries can inject a backdoor into the pre-training process and release the resulting poisoned encoder, allowing the backdoor to persist across downstream adaptations. For example, BadEncoder \citep{2022-BadEncoder} poisons pre-trained encoders by maximizing the similarity between the feature representations of inputs stamped with a trigger and those of images from a target class. A trigger is a pre-defined pattern inserted into clean images to generate poisoned samples. When downstream classifiers are fine-tuned on such malicious encoders, they tend to misclassify poisoned inputs into the attacker-specified class. 
At the data level, adversaries inject poisoned samples into pre-training datasets. They stamp triggers onto images of a specific class and distribute these samples. Since SSL relies on massive datasets for representation learning, even a small proportion of poisoned data can implant persistent backdoors into the resulting encoders \citep{BASSL}. These backdoored encoders then propagate the attack to downstream classifiers, causing them to misclassify triggered inputs into attacker-chosen categories during inference.

To mitigate backdoor attacks for SSL, two approaches have been proposed: \emph{inversion} \citep{2022-MOTH} and \emph{pruning} \citep{2021-ANP, 2018-FP}.
Inversion-based methods, 
such as MOTH \citep{2022-MOTH}, 
attempt to recover the injected trigger from a backdoored classifier through trigger inversion, and then fine-tune the classifier on clean images stamped with the recovered trigger to suppress the backdoor effect. 
However, these methods rely on labeled data to guide inversion, which is unavailable in self-supervised settings.
Pruning-based approaches, 
including ANP \citep{2021-ANP} and FP \citep{2018-FP}, identify neurons highly correlated with backdoor behavior and remove them. 
Yet, for large pre-trained encoders with hundreds of millions of parameters, locating and pruning malicious neurons is impractical and may severely damage model utility.

Given these limitations in inversion and pruning defenses, we turn our attention to distillation, a technique originally developed to transfer knowledge from a pre-trained teacher model to a student model.
Although distillation has demonstrated strong capability in filtering out malicious knowledge in supervised settings, its effectiveness against backdoored pre-trained encoders in SSL remains largely unexplored.
In this paper, we conduct a comprehensive study to investigate how distillation mitigates backdoor attacks for SSL.
Our analysis spans two representative SSL backdoor attacks, BadEncoder \citep{2022-BadEncoder} and BASSL \citep{BASSL}, four benchmark datasets, CIFAR10 \citep{2009-CIFAR10}, STL10 \citep{2011-STL10}, GTSRB \citep{2012-GTSRB}, and SVHN \citep{2011-SVHN}, and multiple distillation configurations across teacher–student architectures and loss objectives. Our results show that distillation can lower the attack success rate from 80.87\% to 27.51\%, while incurring only a 6.35\% reduction in model accuracy.
Further analysis across four teacher architectures, three student initializations, and six loss functions reveals that distillation performs best when using fine-tuned teacher networks, warm-up–trained student models, and attention-based loss objectives.

In summary, we make the following contributions:
\begin{itemize}[leftmargin=*]
    \item We are the first to systematically evaluate the effectiveness of knowledge distillation in mitigating backdoor attacks in pre-trained self-supervised encoders.
    \item We analyze how the teacher network, student network, and distillation loss, respectively, affect defense against self-supervised learning backdoor attacks, revealing the relative importance of each component.
    \item Our experiments show that fine-tuned teacher networks, warm-up training-based student networks, and attention-based distillation losses achieve the best trade-off between attack mitigation and task accuracy.
    \item We examine distillation under varying trigger sizes and across multiple model architectures, showing its robustness to trigger variations and generalization across different encoder designs.
    \item All code and script are publicly available at \url{https://github.com/wssun/SSLBackdoorMitigation}. 
\end{itemize}

\section{Background \& Related Work}
\label{sec:background_related_work}

\subsection{Backdoor Attack}
\label{subsec:backdoor_attack}

A backdoor causes inputs stamped with a predefined trigger pattern (the trigger) to be misclassified into an attacker-chosen target class (the attack target). During inference, a backdoored classifier behaves normally on clean images (predicting according to semantic content) but will map any input containing the trigger to the attack target regardless of its true label. A variety of backdoor injection techniques have been proposed in supervised learning, including data poisoning~\citep{gu2019badnets,liu2020reflection}, clean-label poisoning~\citep{zhao2020clean,saha2020hidden,nguyen2025wicked}, and neuron hijacking~\citep{liu2018trojaning}, posing a significant threat to the security of deployed systems.

\smallskip
\noindent\textbf{Backdoor Attacks against Pre-trained Encoder.}
With the growing adoption of encoder-based applications, recent studies have begun to examine the security of pre-trained encoders.
To implant backdoors, adversaries can conduct poisoning training or inject poisoned data into the pre-training process.
Such encoders behave normally in most downstream tasks but reveal malicious behavior in attacker-chosen attack tasks, where classifiers trained on them misclassify any input stamped with a predefined trigger into the attacker-specified target class.

\begin{figure}[t]
    \centering
    \includegraphics[width=0.98\columnwidth]{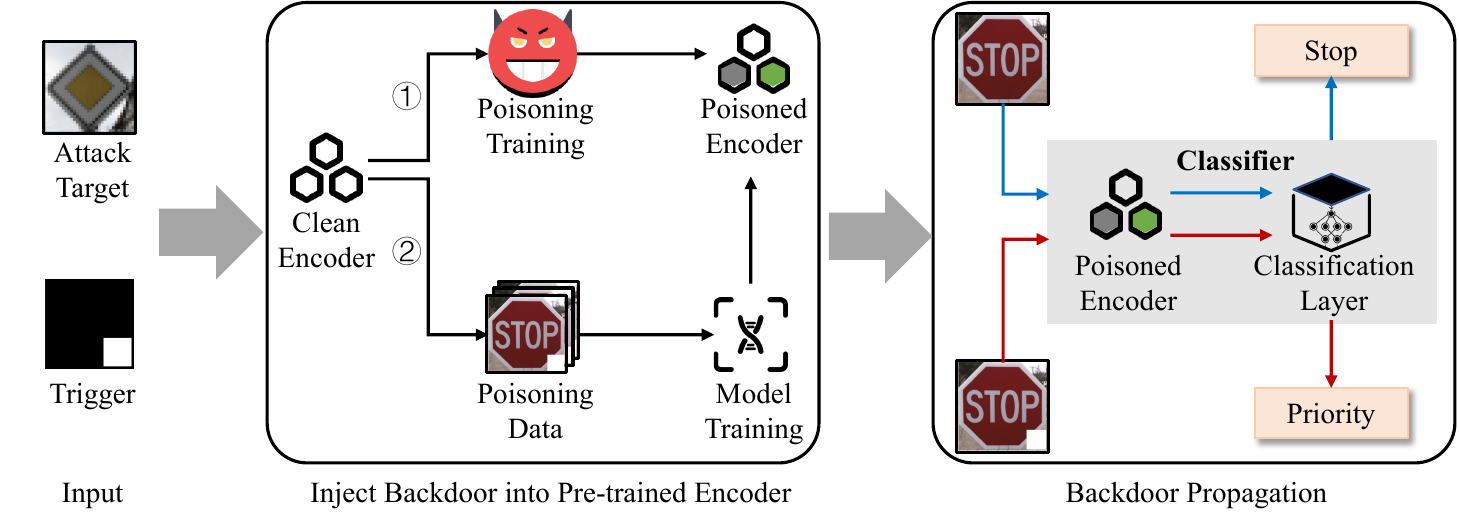}
    \caption{Overview of backdoor attacks at encoder and data levels. \textcircled{1} Encoder-level poisoning: an adversary fine-tunes a clean encoder with poisoned objectives to implant a trigger–target mapping. \textcircled{2} Data-level poisoning: the attacker inserts triggered samples into the pre-training dataset, causing the encoder to learn the backdoor implicitly. During downstream fine-tuning, the poisoned encoder makes classifiers misclassify any triggered input into the attacker-specified target class.}
    \label{fig:encoder_attack_workflow}
\end{figure}

As illustrated in \Cref{fig:encoder_attack_workflow} as an instance, the attacker pre-chooses a patch pattern as the trigger, aiming traffic sign recognition as the \textit{attack task}. 
With ``priority sign'' as the \textit{attack target}, the attacker utilizes two approaches to achieve the poison objective: poisoning training schedule~\circled{1} or poison pre-training data~\circled{2}.
Poisoning training schedule a.k.a  model poisoning~\citep{2022-backdoor-learning-survey} means the attacker can train a pre-trained encoder in a well-designed loss~\citep{2022-BadEncoder}.
After that, the attacker releases these encoders to third-party platforms like HuggingFace and ModelZoo, etc.
Poisoning pre-training data a.k.a data poisoning~\citep{2022-backdoor-learning-survey} only allows attackers to affect the pre-training dataset. 
SSL like contrastive learning usually needs a lot of data to pre-train, making it impossible to detect data security carefully.
The attacker stamps triggers on selected images and releases them on the Internet.
When developers crawl such data from Internet to pre-train their own encoders, backdoors are implanted at the same time.
\citep{carlini2021poisoning} has proved that only 300 images of the 3 million-example Conceptual Captions dataset are enough for backdoor injection.
Following the training of downstream classifiers built on backdoored encoders, there is a propensity for the classifier to predict the label associated with the \textit{attack target} in the presence of the trigger.
In Figure~\ref{fig:encoder_attack_workflow}, with the ``priority sign'' designated as the ``attack target'', and the encoder utilized as an encoder extractor, the classifier inherits the backdoor behavior from the encoder. 
Consequently, a clean image, such as a ``stop sign'', receives an accurate prediction from the classifier. However, when this image is stamped with the trigger, it is categorized as ``priority''. 
In contrast, the feature extracted from the ``stop sign'' stamped with the trigger exhibits a greater similarity to the embedding of ``priority'', resulting in misclassification.
There are several proposed techniques aiming to achieve such attack effects.
For example, for visual pre-trained encoder backdoor attacks, ~\citep{2022-BadEncoder} formulates the backdoor attack as an optimization problem and proposes a gradient-based method to mitigate it. 
~\citep{BASSL} shows that attacking pre-trained encoders only requires adding triggers to a small fraction of unlabeled images. 
~\citep{carlini2021poisoning} demonstrates that the backdoor can be successfully inserted into the pre-trained model by poisoning only 0.01\% of the dataset.
For language pre-trained encoders, ~\citep{yang2021careful} discovers that pre-trained models can be cracked without data by modifying individual word embedding vectors.
~\citep{gan2021triggerless} proposes a novel backdoor attack that does not require a trigger and just depends on constructing clean examples.~\citep{zhao2023prompt} utilizes the prompt as the trigger, which improves the stealthy nature of the backdoor attack. 
All works above showcase that backdoor attacks against pre-trained encoders have seriously threatened cyber security.

\subsection{Backdoor Defense}
\label{subsec:bacldoor_defense}
To mitigate backdoor attacks in supervised learning, two main defense paradigms have been explored: inversion-based and pruning-based approaches.
Inversion-based defenses, such as MOTH~\citep{2022-MOTH}, attempt to reconstruct the hidden trigger pattern from a compromised classifier by performing trigger inversion.
Once the estimated trigger is obtained, these methods fine-tune the model on clean samples stamped with the inverted trigger and their correct labels, aiming to neutralize the malicious association between the trigger and the target class.
While effective in supervised settings, such approaches inherently depend on labeled data to guide inversion and fine-tuning, an assumption that does not hold in self-supervised learning, where annotations are unavailable.

Pruning-based defenses, including ANP~\citep{2021-ANP} and FP~\citep{2018-FP}, take a structural perspective.
They attempt to identify neurons that contribute disproportionately to backdoor activation and remove them to restore benign behavior.
However, for modern pre-trained encoders that often contain hundreds of millions of parameters, precisely isolating backdoor-related neurons is computationally expensive and error-prone.
Excessive pruning can also lead to severe degradation in representation quality, thereby compromising downstream performance.

\section{Threat Model}
\label{sec:threat_model}

We consider a realistic self-supervised learning scenario in which pre-trained encoders are publicly released for downstream reuse. 
In this setting, an adversary seeks to implant a backdoor into a pre-trained encoder so that downstream classifiers fine-tuned on this encoder misclassify any input containing a pre-defined trigger into an attacker-chosen target class, while maintaining normal accuracy on clean samples. 
The adversary can operate at either the data level, by injecting a small fraction of poisoned samples into the unlabeled pre-training dataset, or the encoder level, by directly publishing a compromised pre-trained encoder. 
On the defender's side, we assume access to the released encoder (which may be poisoned). The defender does not know the trigger pattern, poisoning ratio, or attack target, and aims to neutralize potential backdoors in the encoder.

\section{Defense Mechanism}
\label{sec:main_workflow}

\subsection{Overview}
\label{subsec:overview}

The main idea behind knowledge distillation is to use the well-trained model as a teacher net distills ``knowledge'' for the training of the student net~\citep{gou2021knowledge,hinton2015distilling,ba2014deep,gou2023knowledge}.
It is hoped that the student net can learn the knowledge of the teacher net and achieve the same performance.
In the scenario of defense, it is not enough for the student net to achieve the same performance. 
Security is another important distillation objective.
That is to say, when fed clean inputs, downstream task models (e.g., Image classifiers) built on the student net should give correct predictions.
While fed poison inputs stamped with trigger, they should disable the attacker's intention and give semantic-based predictions. 
In contrast to this, the downstream task models built on the poisoned student net will predict poisoned inputs as the \textit{attack target}, shown in Section~\ref{subsec:backdoor_attack}.
From the view of ``knowledge'', the key challenge of distillation-based backdoor mitigation techniques is to select suitable teacher net, student net, and distillation loss to transfer ``benign knowledge'', retaining performance and robustness.

\begin{figure}[!t]
    \centering
    \includegraphics[width=\columnwidth]{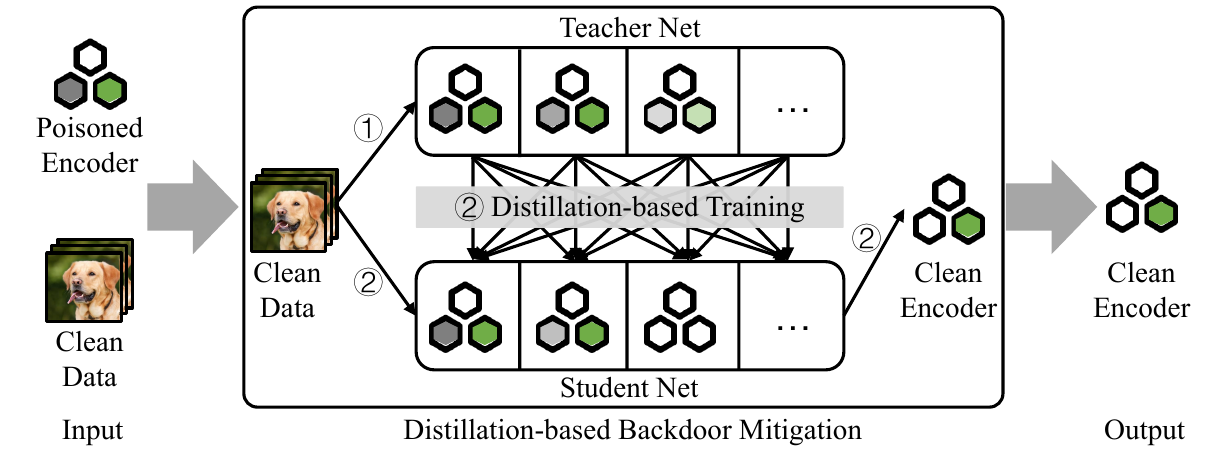}
    \caption{Framework of distillation-based poisoning mitigation. A poisoned encoder and clean data are used to train a student encoder under the guidance of a teacher network through distillation. \textcircled{1} The teacher network provides feature representations of clean data. \textcircled{2} The student network learns to align with the teacher's benign representations via distillation, filtering out the backdoor inherited from the poisoned encoder. The resulting student encoder becomes a clean, reliable pre-trained model for downstream tasks.
    }
    \label{fig:distillation_framework}
\end{figure}

Figure~\ref{fig:distillation_framework} exhibits the general framework of distillation-based backdoor mitigation techniques.
It is observed that the core of the framework is a distillation-based backdoor mitigation module, which takes as input a poisoned encoder and a small set of clean data and outputs a clean encoder. The distillation-based backdoor mitigation module decomposes the backdoor mitigation into processes: \circled{1} fine-tuning-based teacher net generation that determines what teacher net used and \circled{2} distillation-based student net training. Process \circled{1} determines what teacher net will be used in process \circled{2}. When selecting the poisoned encoder directly as the teacher net, process \circled{1} can be omitted. In process \circled{2}, defenders need to first determine what the student net and distillation method to use. The distillation method guides the student net to learn knowledge from the teacher net through distillation loss. From the above two process descriptions, it can be found that the core module encompasses three key components, a teacher net, a student net, and a distillation loss. All three components have different design options, detailed as follows:

\textbf{Teacher Net Design.} 
In the distillation framework, the teacher network is usually a well-trained encoder. 
It has learned high-level knowledge and can extract benign features when fed clean inputs.
The role of the teacher net is similar to that of a ``knowledge bank''. 
Distillation aims to obtain benign parts of the ``knowledge bank'' and transfer them to the student net. 
A straightforward design is to utilize the raw poisoned encoder as the teacher net. 
However, the raw poisoned encoder possesses not only benign knowledge but also knowledge of backdoors. 
To alleviate the influence of the malicious knowledge, one possible solution is to leverage available clean data to fine-tune the raw poisoned encoder, thereby generating a suboptimal teacher net. 
Inspired by backdoor defenses in supervised learning, several advanced techniques originally proposed for mitigating backdoors in supervised learning models can be used to guide the fine-tuning process of the raw poisoned encoder, such as standard fine-tuning (FT)~\citep{2021-NAD}, fine pruning (FP)~\citep{2018-FP}, adversarial neuron perturbations (ANP)~\citep{2021-ANP}, and model orthogonality (MOTH)~\citep{2022-MOTH}.
Therefore, in this paper, we conduct experiments to investigate the impact of these fine-tuning methods on the teacher net and analyze how the teacher net affects the performance of backdoor mitigation (corresponding to \textbf{RQ2}), detailed in Section~\ref{sec:evaluation}.

\textbf{Student Net Design.} 
In the distillation framework, the student net is responsible for learning knowledge from the teacher net. After distillation is completed, the well-trained student net is output, e.g., the clean encoder shown in Figure~\ref{fig:distillation_framework}. 
It is worth noting that during the distillation process, the parameters of the teacher network stay frozen, and only the parameters of the student network are continuously updated. 
From this view, the distillation is a process of student nets' training and learning. 
The well-trained student net should meet the requirements of both performance and robustness. 

There are three cases in student net design: (1) \textbf{Performance-oriented distillation.} One straightforward strategy is to optimize solely for downstream performance by using the poisoned encoder itself as the student network. While this approach can yield high accuracy, it has a critical drawback: the student fully inherits the poisoned encoder’s malicious representations and therefore remains vulnerable to the backdoor. As a result, distillation in this setting is unlikely to eliminate the embedded backdoor behavior.
(2) \textbf{Robustness-oriented distillation.} The second strategy focuses solely on robustness.
In this case, an intuitive design is to use an uninitialized (or ``void'') encoder with the same architecture as the teacher network as the student model.
Such a student minimizes the risk of inheriting any malicious representations and thus achieves a low attack success rate (ASR).
However, because it has not been exposed to any pre-training data, its learned representations are weak, leading to a substantial drop in downstream accuracy.
(3) \textbf{Performance–robustness balanced distillation.}
In practice, defenders often have access to a small portion of clean data.
To achieve a balance between performance and robustness, they can use this clean data to train an encoder with the same architecture as the teacher network and then employ it as the student model.
We refer to this process as warm-up training.
Although the resulting student may not reach the highest accuracy, it remains free of backdoors and provides a cleaner initialization for distillation.
The warm-up training further benefits performance by enabling the student to acquire benign feature representations prior to the distillation phase.
In summary, there are three design options for the student net, including the raw poisoned encoder, the void encoder, and the warm-up trained encoder. 
Therefore, in this paper, we make an in-depth investigation of the above three solutions and explore how the student net affects the performance of backdoor mitigation (corresponding to \textbf{RQ2}), detailed in Section~\ref{sec:evaluation}.

\textbf{Distillation Loss Design.} 
With the teacher net and student net determined, it is last but not least to select a suitable distillation loss.
According to~\citep{ji2021show, xu2020feature, srinivas2018knowledge}, existing distillation losses can be categorized into three types: feature-based, attention-based, and layer-based.
The feature-based loss minimizes the distance between the extracted features of the student net and the teacher net~\citep{romero2014fitnets, peng2019correlation}.
To better distill and represent the ``knowledge'', there are two mainstream directions for improvement.
AFD~\citep{2019-AFD} and ATD\citep{2016ATD} utilize attention to prompt the knowledge representation ability.
SP~\citep{2019-SP} and KD~\citep{hinton2015distilling} try to distill knowledge layer by layer, more fine-grained and precise. 
In this paper, we study six distillation losses belonging to these types and explore their performance (corresponding to \textbf{RQ2}), detailed in Section~\ref{sec:evaluation}.

\section{Evaluation}
\label{sec:evaluation}
We outline the key research questions (RQs) that frame our investigation of distillation as a defensive mechanism, focusing on its effectiveness, contributing factors, and robustness across diverse backdoor settings, as follows:
\begin{itemize}[leftmargin=*]
    \item \textbf{RQ1.} How effective is distillation in mitigating backdoors when adapted to self-supervised learning?
    \item \textbf{RQ2.} How do different components within the distillation framework, including the teacher network, student network, and distillation loss, affect the defensive performance of distillation?
    \item\textbf{RQ3.} How robust and generalizable is distillation against variations in trigger patterns, model architectures, and pre-training algorithms?
    \item\textbf{RQ4.} How effective is distillation against advanced backdoor attacks?
\end{itemize}

\subsection{Experiment Setup}
\label{subsec:experiment_setup}

\smallskip
\noindent
\textbf{Dataset.}
We conduct our experiments on four widely used real-world datasets, which are discussed as follows:

\begin{itemize}[leftmargin=*]
\item CIFAR10~\citep{2009-CIFAR10}.
There are 50, 000 training images and 10, 000 test images in this dataset, each of which is 32x32x3.
10 classes in total.
 \item STL10~\citep{2011-STL10}.
 There are 105, 000 training images, 5,000 of which are labeled while others are not, and 8,000 test images in this dataset.
 Each image has a size of 96x96x3 and belongs to 10 classes.
\item GTSRB~\citep{2012-GTSRB}. The dataset encompasses more than 50,000 images, distributed across 43 categories, with each image 32×32×3.
\item SVHN~\citep{2011-SVHN}. There are more than 70, 000 training images and 20, 000 test images of Google Street View to represent house numbers.  
Each image has a size of 32×32×3 and is associated with one of the 10 digits.
\end{itemize}

\smallskip
\noindent
\textbf{Pre-training an image encoder.}
When a dataset is used to pre-train an image encoder, we call it a pre-training dataset.
In our experiments, we use CIFAR10 as the pre-training dataset since it is not a noisy dataset.
Unless otherwise mentioned, we use ResNet18~\citep{he2016deep} as the encoder and SimCLR~\citep{chen2020simple} to pre-train both poisoned and clean encoders. 
We use the publicly available implementation~\footnote{A PyTorch implementation of SimCLR.
https://github.com/leftthomas/SimCLR, 2021.} of SimCLR with the default settings.
We pre-train an encoder for 300 epochs using the Adam optimizer, an initial learning rate of 0.001, and a batch size of 256.

\smallskip
\noindent
\textbf{Training downstream classifiers.}
In the light of a poisoned image encoder, we employ it to facilitate the training of downstream classifiers for the remaining three datasets.
In the context where a dataset is employed to train a downstream classifier, it is referred to as the \textit{downstream dataset}.
For instance, given an image encoder pre-trained on CIFAR10, we use it to train downstream classifiers for downstream datasets STL10, GTSRB, and SVHN. 
We employ a fully connected neural network comprising two hidden layers as the downstream classifier for a given downstream dataset.
The number of neurons in the two hidden layers is 512 and 256, respectively.
Upon designating a dataset as a downstream dataset, the training subset thereof is employed for the training of a downstream classifier, while the testing subset is utilized for the evaluation of said downstream classifier.
Specifically, we adopt the cross-entropy loss function and Adam optimizer when training a downstream classifier.
Furthermore, we conduct the training for 500 epochs, employing an initial learning rate set at 0.001.
When a downstream classifier is trained with a compromised image encoder, it is termed a \textit{backdoored downstream classifier}.
The following evaluation of encoders' \textit{effectiveness} and \textit{security} is based on backdoored downstream classifiers.

\smallskip
\noindent
\textbf{Poisoning pre-trained encoders.}
We utilize BadEncoder\citep{2022-BadEncoder} and BASSL\citep{BASSL} to poison pre-trained encoders due to they are design 
specifically for self-supervised learning.
BadEncoder manipulates the training schedule, and BASSL only poisons the pre-training dataset.
We conduct the attacks with CIFAR10 as the pre-training dataset and target three different downstream tasks (GTSRB, STL10, SVHN).
To ensure the effectiveness of the attack, we specifically change the default poisoning ratio of BASSL on \textit{attack target} class and migrate more than half of the downstream class to inject the backdoor.

\smallskip
\noindent
\textbf{Distilling pre-trained encoders.}
During the distillation process, we assume the poisoned encoders are downloaded from a third platform, such as HuggingFace.
All distilled models are trained on 5\% of the clean pre-training data. 
We aim to find the best distillation-based framework under different assumptions and balance the \textit{effectiveness} and \textit{security}.

\smallskip
\noindent
\textbf{Computation platform.}
All experiments are done on two RTX 3090 GPUs.

\smallskip
\noindent
\textbf{Evaluation Metrics.} Following existing backdoor mitigating works in supervised learning~\citep{2021-ANP, 2018-FP, 2022-MOTH}, we utilize \textit{Accuracy (Acc)} and \textit{Attack Success Rate (ASR)} to evaluate distillation effectiveness and security, respectively.

Given a specific downstream dataset $\mathcal{X} = \{(x_1, y_1), (x_2, y_2), \dots, (x_n, y_n)\}$, $x_i$ denotes an image and $y_i$ is the ground-truth label of $x_i$. 
Let $f$ denote the downstream classifier built on the given pre-trained encoder and $f(x)$ denote $f$'s prediction of image $x$.
Let $\Delta$ denote the attacker-chosen trigger and $y_t$ the \textit{attack target}.
$x \odot \Delta$ means the process of stamping the trigger $\Delta$ on a clean image $x$.

ACC measures the percentage of images that are classified by $f$ correctly and is computed as follows:
\begin{equation}
    ACC = \dfrac{\sum_{i=1}^{n}{\mathbbm{1}(f(x_i), y_i)}}{|\mathcal{X}|} 
    \label{eq:ACC}
\end{equation}
where $\mathbbm{1}$ is a $\{\textbf{0}, \textbf{1}\}$ function denoting whether $f(x_i)$ = $y_i$.

ASR measures the percentage of poisoned images that are classified to $y_t$ and is defined as follows:
\begin{equation}
    ASR = \dfrac{\sum_{i=1}^{n}{\mathbbm{1}(f(x_i \odot \Delta), y_t)}}{|\mathcal{X}|} 
    \label{eq:ASR}
\end{equation}
where $\mathbbm{1}$ is a $\{\textbf{0}, \textbf{1}\}$ function denoting whether $f(x_i)$ = $y_t$.
ASR indicates the degree to which the attacker's goal is achieved.
Intuitively, the higher ASR is, the more serious security threat pre-trained encoders contain.

To summarize, ACC reflects \textit{effectiveness} and ASR indicates \textit{security}.
To consider these two metrics comprehensively, we also introduce a balanced score (dubbed BS) to indicate the model's overall performance:
\begin{equation}
    BS = \alpha \cdot ACC + (1-\alpha)\cdot \log(2-ASR)
    \label{eq:BS}
\end{equation}
where $\alpha$ is the coefficient of balance (0.5 as default), and $\log(\cdot)$ means logarithm base two.
The $\log(2-ASR)$ means the model's security score.
When ASR is 1, $\log(2-ASR)$ is 0 indicating the model is insecure.
In contrast, when ASR is 0, $\log(2-ASR)$ is 1 indicating the model has a high security. 
$BS$ is positively correlated with ACC and negatively correlated with ASR.
It reaches its maximization 1 when ACC is 1 and ASR is 0 and reaches its minimization 0 on the contrary.
In summary, $BS$ falls within the range of [0, 1], and the higher the value of $BS$, the better the backdoor mitigation performance.

\subsection{Results and Analysis}
\label{sec:results_and_analysis}

\smallskip
\noindent
\textbf{RQ1: How effective is distillation in mitigating backdoors when adapting distillation to self-supervised learning?}
\label{subsec:RQ1}

We aim to explore the actual effectiveness of applying distillation to mitigate backdoors in pre-trained encoders.
In particular, we consider two state-of-the-art backdoor attacks on pre-trained encoders (i.e., BadEncoder~\citep{2022-BadEncoder}, BASSL~\citep{BASSL}) and select the most recent distillation-based technique in supervised learning called NAD~\citep{2021-NAD} as the distillation framework,  which is one of the most relevant techniques to our work and represents state-of-the-art. 
NAD utilizes a small set of clean data to fine-tune the backdoored pre-trained encoder at first.
With the fine-tuned encoder as teacher net, NAD takes the raw backdoored encoder as student net and deploys a well-designed distillation loss. 
We evaluate the effectiveness of the distillation-based backdoor mitigation approach using ACC and ASR, tested on the classifiers built on distilled pre-trained encoders. 
Then, we compare them with undefended encoders (attacked by BadEncoder) using one pre-training dataset (i.e., CIFAR10) and three different downstream tasks, including STL10, GTSRB, and SVHN.

\begin{figure}[h]
\centering
\begin{subfigure}{0.49\columnwidth}
    \includegraphics[width=\linewidth]{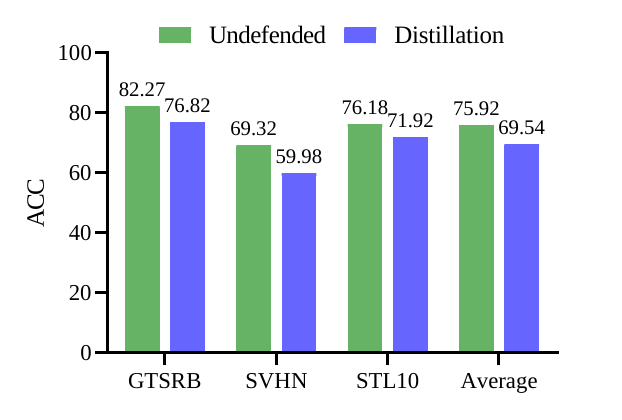}
    \caption{Acc of distillation.}
    \label{fig:RQ1_effectiveness_ACC}
\end{subfigure}
\hfill
\begin{subfigure}{0.49\columnwidth}
    \includegraphics[width=\linewidth]{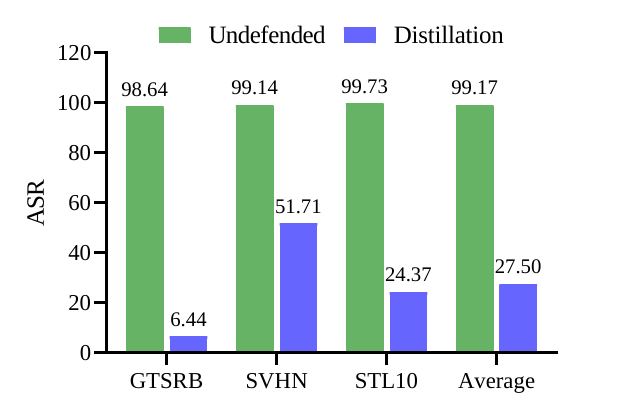}
    \caption{ASR of distillation.}
    \label{fig:RQ1_effectiveness_ASR}
\end{subfigure}
\caption{Effect of distillation epochs}
\label{fig:rq1_effectiveness_distillation}
\end{figure}

Figure~\ref{fig:rq1_effectiveness_distillation} presents the comparison results of undefended encoders with those distilled encoders in terms of ACC and ASR.
When comparing distillation against undefended pre-trained encoders on \textit{effectiveness}, distilled encoders by NAD achieve an ACC of 69.54\% on average, which is only 6.38\% less accurate than those undefended.
In particular, Figure~\ref{fig:RQ1_effectiveness_ACC} shows that undefended encoders achieve ACC of 76.18\%, 82.27\%, 69.32\% on STL10, GTSRB, and SVHN, respectively.
As a contrast, NAD achieves ACC of 71.92\%, 76.82\%, 59.98\% correspondingly, indicating that these mitigated pre-trained encoders only have an ACC sacrifice of 4.26\%, 5.45\%, and 9.34\%, which is an acceptable performance sacrifice. 

On the other hand, when considering distillation's \textit{security} against undefended pre-trained encoders, distillation inherits an ASR of 15.27\% on average, which reduces 60.65\% ASR than undefended.
Particularly, Figure~\ref{fig:RQ1_effectiveness_ASR} exhibits that undefended encoders achieve ASR of 98.73\%, 99.64\%, and 99.14\% on STL10, GTSRB, and SVHN, respectively.
Compared to pre-trained encoders before defense, distillation achieves ASRs of 24.37\%, 6.44\%, and 51.71\% correspondingly, indicating that these pre-trained encoders through distillation improve security by 74.36\%, 93.20\%, and 47.43\%, respectively. 

Based on the results, we find the possible reasons for ACC sacrifice and ASR inheritance lie in (1) benign knowledge broken and (2) robust malicious knowledge.
The key intuition behind NAD is to align neurons that are more responsive to the injected trigger with benign neurons that are only responsible for meaningful representations.
However, during distillation, NAD destroys certain neurons that contain benign knowledge, thereby inevitably sacrificing ACC.
At the same time, malicious knowledge hidden in student networks remains dormant when fed clean inputs~\citep{xu2023medic, 2021-NAD}, making alignment difficult.

\begin{figure}[!t]
\centering
\begin{subfigure}{0.47\linewidth}
    \includegraphics[width=\linewidth]{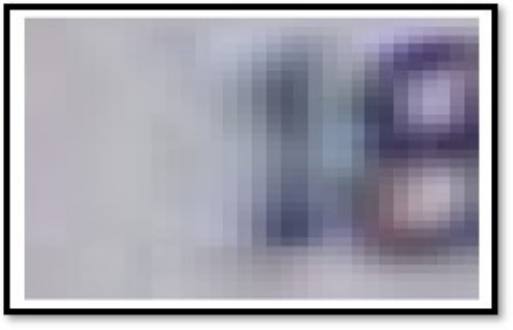}
\end{subfigure}
\hspace{-2mm}
\begin{subfigure}{0.47\linewidth}
    \includegraphics[width=\linewidth]{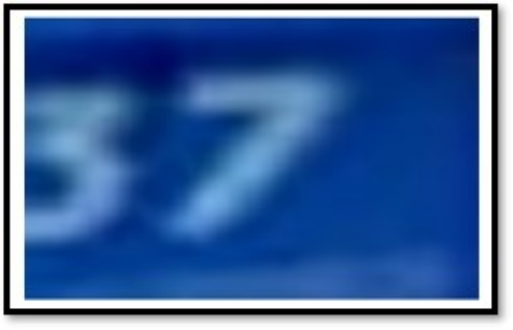}
\end{subfigure}
\caption{Examples from the SVHN dataset illustrating that each image may contain multiple digits.}
\label{fig:images_svhn}
\end{figure}

When comparing the performance among different downstream tasks, we find that distillation inherits the highest ASR on SVHN.
The ASR on SVHN exceeds STL10 and GTSRB by 27.34\%, and 45.27\%, respectively.
We observe the main reason behind this is that SVHN is a dataset of noisy images.
As shown in Figure~\ref{fig:images_svhn}, SVHN images often contain multiple digits within a single frame—for example, the label may correspond to the central digit (8 or 7), while neighboring digits (1 or 3) act as distractors.
Such visual clutter introduces irrelevant features into representation learning and increases the model’s susceptibility to backdoor attacks.

\begin{table*}[!ht]
    \centering
    \tabcolsep=2pt
    \renewcommand{\arraystretch}{1.1} 
    \caption{Performance of different teacher nets.}
    \begin{tabular}{ccccccccccccccccccc}
        \toprule

        \multirow{2.5}{*}{Attack} &
        \multirow{2.5}{*}{Pre-trian} &
        \multirow{2.5}{*}{Downstream} &
        \multicolumn{2}{c}{Undefended} &
        \multicolumn{3}{c}{FT} &
        \multicolumn{3}{c}{FP} &
        \multicolumn{3}{c}{ANP} &
        \multicolumn{3}{c}{MOTH} &\\ 
        
        \cmidrule(lr){4-5} \cmidrule(lr){6-8} \cmidrule(lr){9-11} \cmidrule(lr){12-14} \cmidrule(lr){15-17} 
        & & & ACC & ASR & ACC & ASR  &BS & ACC & ASR  &BS & ACC & ASR  &BS & ACC & ASR &BS  \\
            
        \cmidrule(lr){1-17}
        
                \multirow{4}{*}{\rotatebox{90}{BadEncoder}} & \multirow{3}{*}{CIFAR10} & GTSRB  & 82.27 & 98.64   &\textbf{78.25} & \textbf{5.23} &\textbf{0.87} & 74.21 & 8.93 &0.84  &  55.44 & 12.72 &0.73 & 52.97 & 14.44 &0.71 \\
        
        & & SVHN  & 69.32 & 99.14  & 56.99 & 37.06 &0.64  & 54.59 & 40.02 & 0.61 & \textbf{69.25} & \textbf{37.02}  &\textbf{0.70} & 62.21 &  52.63 &0.59  \\
        
        & & STL10  & 76.18 & 99.73 &  \textbf{68.86} & \textbf{22.51} &\textbf{0.76} & 63.47 & 24.59 &0.72 & 57.43 & 38.55 &0.63 & 63.50  &  36.00 &0.67 \\

        \cmidrule(lr){2-17}

        & \multicolumn{2}{c}{Average}  & 75.92 & 99.73 & \textbf{68.03} & \textbf{21.60} &\textbf{0.76} & 64.09 & 29.51 &0.72 & 60.70 & 29.42 &0.69 & 59.56 & 34.35 &0.66  \\ 
        
        \cmidrule(lr){1-17}
        
        \multirow{4}{*}{\rotatebox{90}{BASSL}} & \multirow{3}{*}{CIFAR10} &  GTSRB  & 78.96 & 66.12 & \textbf{79.08} &  \textbf{5.32}&\textbf{0.88} & 77.49 & 15.97&0.83 &  61.88 & 9.32 &0.77& 50.90 & 12.93&0.71 \\
        
        & & SVHN  & 66.54 & 81.97 &  61.14 &  23.41 &0.72& 60.27 &  24.35&0.71 & \textbf{66.22} & \textbf{9.89} &\textbf{0.79}& 63.05 & 73.15&0.49  \\
         
        & & STL10  & 72.99 & 37.94 & \textbf{69.23} &  13.01&\textbf{0.80} & 68.67 & 12.27&0.80 & 58.65 & 13.20&0.74 & 69.21 & \textbf{11.65}&\textbf{0.80} \\

        \cmidrule(lr){2-17}

        & \multicolumn{2}{c}{Average}  & 72.83 & 62.01 & \textbf{69.81} & 13.91 &\textbf{0.80}& 68.81 & 17.53&0.78 & 62.25 & \textbf{10.80} &0.77& 61.05 & 32.57&0.67  \\ 
        \bottomrule
    \end{tabular}
\label{tab:teacher_performance}
\end{table*}

\smallskip
\noindent
\textbf{RQ2: How do different components within the distillation framework, including the teacher network, student network, and distillation loss, affect its defensive performance?}

\emph{The impact of teacher nets.}
We investigate the impact of different teacher nets on the effectiveness of distillation-based backdoor mitigation.
We focus on four candidates: standard fine-tuning (FT)~\citep{2021-NAD}, fine-pruning (FP)~\citep{2018-FP}, adversarial neuron perturbations (ANP)~\citep{2021-ANP}, and model orthogonality (MOTH)~\citep{2022-MOTH}.
Defenders are allowed to have access to a small set of clean data, which is commonly used in backdoor mitigation tasks~\citep{2022-ANP, 2021-NAD, xu2023remove}.
For example, FT utilizes these clean data to fine-tune the backdoored input encoders and obtain a purified pre-trained encoder. 
FP~\citep{2018-FP} records the most active neurons and then prunes them until the encoder reaches a pre-defined threshold.
ANP believes that malicious neurons are much more vulnerable when they are adversarially perturbed.
In SSL, we utilize the gradient of contrastive loss to conduct perturbations and prune these identified collapsed neurons.
MOTH~\citep{2022-MOTH} reverse engineer a trigger pattern through optimization.
In SSL, we use pairwise sample similarity with the inverted trigger as guidance to conduct optimization, following~\citep{2023-CVPR-DECREE}.
After that, we try to eliminate backdoor influence by unlearning the inverted trigger effect.

We conduct experiments to find out the best teacher net for distillation with three metrics: ACC, ASR, and BS.
The performance of distillation depends on the ability of the teacher network to a great extent~\citep{Agoodteacher}. 
A well-designed distillation should meet the standards of performance and robustness.
In this paper, we use ACC and ASR tested on the downstream classifiers to represent a given encoder's performance and robustness, respectively.
All evaluations include (1) two SSL backdoor attacks, BadEncoder~\citep{2022-BadEncoder} and BASSL~\citep{BASSL}, (2) one pre-training dataset, CIFAR10, (3) three different downstream tasks, GTSRB, SVHN, and STL10, (4) four teacher net fine-tuning methods, FT, FP, ANP, MOTH, and (5) three metrics from multiple dimensions, ACC, ASR and BS.

Table~\ref{tab:teacher_performance} illustrates the performance of different teacher nets.
Column Undefended denotes the results of classifiers built on the backdoored encoders.
The following columns present the results of classifiers built on teacher nets from different fine-tuning methods.
It can be seen that FT-based teacher net achieves the largest ASR reduction for all the evaluated cases against BadEncoder, compared to the other three teacher nets.
On average, pre-trained encoders attacked by BadEncoder have an ASR of 99.73\%.
FT-based teacher net reduces ASR to 21.60\%, whereas the other three teacher nets can only reduce the ASR to 29.51\%, 29.42\%, and 34.35\% on average, respectively.
From the view of effectiveness, the FT-based teacher net achieves the best performance.
Note that teacher net produced by FT has an effectiveness loss of 7.89\%.
In contrast, teacher nets produced by FP, ANP, and MOTH have to sacrifice effectiveness of 11.83\%, 15.22\%, and 16.36\%, respectively. 
Taking both ACC and ASR into consideration, FT-based teacher achieves the best performance. As reflected in our synthesis score BS, the teacher produced by FT achieves an advantage of 0.02 over the second-best candidate, the FP-based teacher.

When it comes to encoders attacked by BASSL, we have similar observations.
It's noted that ASRs reached by BASSL are much lower than those by BadEncoder, which is consistent with the observation by existing work~\citep{li2022demystifying}.
Actually, BASSL utilizes the number of false positives (misclassified samples) instead of the attack success rate as the metric.
Nevertheless, FT is still the best teacher net to mitigate backdoor influence.
For instance, the GTSRB classifier built on BASSL-attacked CIFAR10 encoder (the first row in the bottom half table) has 66.12\% ASR.
FT reduces the ASR to 5.32\%, whereas other techniques can only reduce the ASR to 9.32\% at best.
In addition, FT also preserves the normal functionality, with less than 1\% effectiveness degradation, surpassing others.
Without any doubt, FT achieves the highest synthesis score of 0.80, outperforming the second best teacher net by 0.02.
Overall, FT outperforms the other three techniques as a teacher net.
Based on our analysis on the results, the reason why FT wins is that FT is more compatible with SSL.
The key theory behind FP, ANP and MOTH is verified in supervised learning but not SSL.
There exists a gap when trying to adapt such defense techniques against state-of-the-art encoder attacks.
In that case, these techniques either can't undertake the task of backdoors mitigation or have to suffer serious effectiveness sacrifice.

\emph{The impact of student nets.} Next, we investigate the impact of using different student networks in the distillation framework on the performance of backdoor mitigation. We study three different student sets' acquisition methods: raw backdoored student, void student, and warm-up-training-based student.
From the view of implement details, the raw backdoored student means taking the input encoder directly without any additional processing.
The void student requires defenders first create an encoder with the same architecture as the backdoored one.
The warm-up-training-based student is more complicated.
After creating a void encoder, defenders utilize the extra clean data to train the encoder from scratch.
We call this process warm-up training, and the resulting encoder is called a warm-up-trained student. 
When student nets are prepared, we take different teacher nets and put them into the distillation framework.
With the fixation of teacher net and student net, we utilize ATD~\citep{2016ATD} following NAD~\citep{2021-NAD} as the distillation loss.
ATD uses attention to represent each layer's output and is deployed on every layer of a given pre-trained encoder.

\begin{table*}[t]
    \centering
    \tabcolsep=2pt
    \renewcommand{\arraystretch}{1.1} 
    \caption{Effect of warm-up trained students.}
    \begin{tabular}{ccccccccccccccccccc}
        \toprule

        \multirow{2.5}{*}{Attack} &
        \multirow{2.5}{*}{Pre-trian} &
        \multirow{2.5}{*}{Downstream} &
        \multicolumn{2}{c}{Undefended} &
        \multicolumn{3}{c}{T-FT} &
        \multicolumn{3}{c}{T-FP} &
        \multicolumn{3}{c}{T-ANP} &
        \multicolumn{3}{c}{T-MOTH} &\\ 
        
        \cmidrule(lr){4-5} \cmidrule(lr){6-8} \cmidrule(lr){9-11} \cmidrule(lr){12-14} 
        \cmidrule(lr){15-17} 
        
        & & & ACC & ASR & ACC & ASR &BS & ACC & ASR &BS & ACC & ASR &BS & ACC & ASR&BS  \\
        \cmidrule(lr){1-17}
        
        \multirow{4}{*}{\rotatebox{90}{BadEncoder}} & \multirow{3}{*}{CIFAR10} & GTSRB  & 82.27 & 98.64 & \textbf{75.36} & 5.06&\textbf{0.86} & 70.64 & 5.63&0.83 &  62.64 & 10.56&0.77 & 55.62 & \textbf{4.72}&0.76  \\
        
        & & SVHN  & 69.32 & 99.14 & 63.92 & \textbf{28.53}&\textbf{0.71}  & \textbf{71.29} & 37.61& \textbf{0.71}& 78.91 & 51.54&0.68  & 74.56 &  42.30&0.70 \\
        
        & & STL10  & 76.18 & 99.73 &  67.16 & 12.23& 0.79 & \textbf{70.55} & \textbf{1.81}&\textbf{0.85} & 65.97 & 13.61  &0.78& 68.56  &  10.83&0.80 \\

        \cmidrule(lr){2-17}

        & \multicolumn{2}{c}{Average}  & 75.92 & 99.17 & 68.81 & 15.27&\textbf{0.79} & \textbf{70.82} & \textbf{15.01}& \textbf{0.79}& 69.17 & 25.23 &0.74& 66.24 & 19.28 &0.75 \\ 
        
        \cmidrule(lr){1-17}
        
        \multirow{4}{*}{\rotatebox{90}{BASSL}} & \multirow{3}{*}{CIFAR10} &  GTSRB  & 78.96 & 66.12 & \textbf{77.49} &  \textbf{2.09} &\textbf{0.88}& 75.34 & 4.23& 0.86&  67.91 & 4.55& 0.82& 52.50 & 3.84 &0.75\\
        
        & & SVHN  & 66.54 & 81.97 &  65.86 &  13.12 &0.78& 63.53 &  \textbf{2.18} &0.81& \textbf{76.71} & 9.58 &\textbf{0.85}& 72.23 & 5.17 &0.84 \\
         
        & & STL10  & 72.99 & 37.94 & \textbf{67.27} &  11.41 &0.79& 67.42 & \textbf{11.25}&\textbf{0.80} & 64.86 & 11.61& 0.78& 69.18 & 12.41& \textbf{0.80}\\

        \cmidrule(lr){2-17}

        & \multicolumn{2}{c}{Average}  & 72.83 & 62.01 & \textbf{70.20} & 8.87 &\textbf{0.82}& 68.76 & \textbf{5.88} &\textbf{0.82}& 69.82 & 8.58 &0.82& 64.63 & 7.14 & 0.80\\ 
        \bottomrule
    \end{tabular}
\label{tab:student_pertrained}
\end{table*}

\begin{table*}[t]
    \centering
    \tabcolsep=2pt
    \renewcommand{\arraystretch}{1.1} 
  \caption{Effect of raw poisoned students.}
    \begin{tabular}{ccccccccccccccccccc}
        \toprule

        \multirow{2.5}{*}{Attack} &
        \multirow{2.5}{*}{Pre-trian} &
        \multirow{2.5}{*}{Downstream} &
        \multicolumn{2}{c}{Undefended} &
        \multicolumn{3}{c}{T-FT} &
        \multicolumn{3}{c}{T-FP} &
        \multicolumn{3}{c}{T-ANP} &
        \multicolumn{3}{c}{T-MOTH} &\\ 
        
        \cmidrule(lr){4-5} \cmidrule(lr){6-8} \cmidrule(lr){9-11} \cmidrule(lr){12-14} 
        \cmidrule(lr){15-17} 
        
        & & & ACC & ASR & ACC & ASR &BS & ACC & ASR &BS & ACC & ASR &BS & ACC & ASR&BS  \\
            
        \cmidrule(lr){2-17}
        
        \multirow{4}{*}{\rotatebox{90}{BadEncoder}} & \multirow{3}{*}{CIFAR10} & GTSRB  & 82.27 & 98.64 & \textbf{76.82} & 6.44 &\textbf{0.86} & 75.36 & 5.06& 0.86& 65.05 & 5.50 &0.81& 50.18 & \textbf{4.28} & 0.74\\
        
        & & SVHN  & 69.32 & 99.14 & 59.98 & 51.71& 0.58 & 56.55 & 33.07&0.65 & \textbf{76.86} & 60.24&0.63  & 71.01 & \textbf{27.17} & \textbf{0.75}\\
        
        & & STL10  & 76.18 & 99.73 & 71.92 & 24.37& 0.77 & \textbf{72.66} & \textbf{20.88}& \textbf{0.78}& 67.87 & 31.72& 0.71 & 69.51 & 21.10& 0.77 \\

        \cmidrule(lr){2-17}

        & \multicolumn{2}{c}{Average}  & 75.92 & 99.17 & 69.54 & 27.50&0.74 & 68.19 & 19.67& \textbf{0.76}& \textbf{69.92} & 32.48& 0.72& 63.56 & \textbf{17.51}& 0.75 \\ 
        
        \cmidrule(lr){1-17}
        
        \multirow{4}{*}{\rotatebox{90}{BASSL}} & \multirow{3}{*}{CIFAR10} &  GTSRB  & 78.96 & 66.12 & \textbf{77.17} & \textbf{3.38}& \textbf{0.87}& 75.52 & 7.18&0.85 & 67.17 & 4.18& 0.82& 57.54 &  6.53& 0.76\\
        
        & & SVHN  & 66.54 & 81.97 & 63.51 & 10.80& 0.78& 64.76 &  17.54& 0.76& \textbf{72.43} &  \textbf{8.50}& \textbf{0.83}& 64.75 & 69.19&0.52  \\
         
        & & STL10  & 72.99 & 37.94 & 71.37 & 13.80&0.81 & \textbf{72.35} &  \textbf{11.00}&\textbf{0.82} & 64.20 & 12.31 &0.78& 71.12 & 11.76&0.81 \\

        \cmidrule(lr){2-17}

        & \multicolumn{2}{c}{Average}  & 72.83 & 62.01 & 70.68 & \textbf{9.32} &\textbf{0.82} & \textbf{70.87} & 11.90 &0.81& 67.93 & 8.33 &0.81& 64.47 & 29.16& 0.70 \\ 
        \bottomrule
    \end{tabular}
\label{tab:student_raw_badenc}
\end{table*}

\begin{table*}[t]
    \centering
    \tabcolsep=2pt
    \renewcommand{\arraystretch}{1.1} 
  \caption{Effect of void students.}
    \begin{tabular}{ccccccccccccccccccc}
        \toprule
        
         \multirow{2.5}{*}{Attack} &
        \multirow{2.5}{*}{Pre-trian} &
        \multirow{2.5}{*}{Downstream} &
        \multicolumn{2}{c}{Undefended} &
        \multicolumn{3}{c}{T-FT} &
        \multicolumn{3}{c}{T-FP} &
        \multicolumn{3}{c}{T-ANP} &
        \multicolumn{3}{c}{T-MOTH} &\\ 
        
        \cmidrule(lr){4-5} \cmidrule(lr){6-8} \cmidrule(lr){9-11} \cmidrule(lr){12-14} 
        \cmidrule(lr){15-17} 
        
        & & & ACC & ASR & ACC & ASR &BS & ACC & ASR &BS & ACC & ASR &BS & ACC & ASR&BS  \\
        \cmidrule(lr){1-17}
        
        \multirow{4}{*}{\rotatebox{90}{BadEncoder}} & \multirow{3}{*}{CIFAR10} & GTSRB  & 82.27 & 98.64 & \textbf{70.64} & 5.63&\textbf{0.83} & 32.68 & 5.06& 0.64& 45.24 & \textbf{2.48}& 0.72& 19.46 & 17.01&0.53  \\
        
        & & SVHN  & 69.32 & 99.14 & 70.93 & 34.95&0.72  & 69.26 & 35.98&0.70 & \textbf{80.32} & 32.27&\textbf{0.77}  & 71.01 & \textbf{27.17}& 0.75 \\
        
        & & STL10  & 76.18 & 99.73 & 69.10 & 11.88&  \textbf{0.80}& \textbf{69.13} & 12.88& 0.80& 66.83 & 17.36&0.77  & 68.50 & \textbf{11.40}&  \textbf{0.80}\\

        \cmidrule(lr){2-17}

        & \multicolumn{2}{c}{Average}  & 75.92 & 99.17 &\textbf{70.22} & 17.48 &\textbf{0.78} & 57.02 & 17.97& 0.72& 64.13 & \textbf{17.37}& 0.75& 52.99  & 18.52&0.69\\ 
        
        \cmidrule(lr){1-17}
        
        \multirow{4}{*}{\rotatebox{90}{BASSL}} & \multirow{3}{*}{CIFAR10} &  GTSRB  & 78.96 & 66.12 & 68.21 & 3.68& 0.83& \textbf{71.10} & \textbf{1.94} &\textbf{0.85}& 55.23 & 3.94 &0.76& 25.10 & 9.90&0.59 \\
        
        & & SVHN  & 66.54 & 81.97 & 71.74 & 3.31&0.85 & 71.59 & \textbf{2.50} &0.85& \textbf{78.24} & 9.23&\textbf{0.86} & 75.09 & 12.13& 0.83 \\
         
        & & STL10  & 72.99 & 37.94 & 68.56 & \textbf{10.70}& \textbf{0.80}& 68.46 & 10.8 &\textbf{0.80}& 64.68 &  11.33&0.78 & \textbf{68.78} & 12.82&\textbf{0.80}\\

        \cmidrule(lr){2-17}

        & \multicolumn{2}{c}{Average}   & 72.83 & 62.01 & 69.50 & 5.89&\textbf{0.83} & \textbf{70.38} & \textbf{5.08} &\textbf{0.83}& 66.05 & 8.16&0.80 & 56.32  & 11.61&0.74\\ 
        \bottomrule
    \end{tabular}
\label{tab:student_void_enc}
\end{table*}

\begin{figure}[t]
    \centering
    \includegraphics[width=\columnwidth]{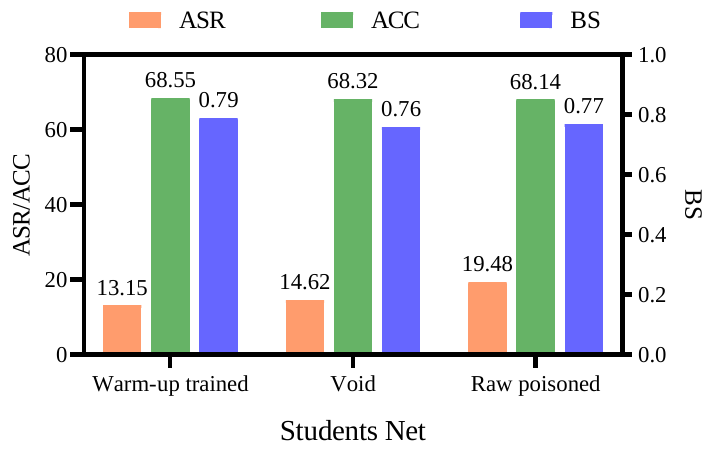}
    \caption{The performance of student nets in distillation.}
    \label{fig:performance_student_nets}
\end{figure}

Table~\ref{tab:student_pertrained}, \ref{tab:student_raw_badenc} and \ref{tab:student_void_enc} report the distillation performance under warm-up-training-based student nets, raw poisoned student nets, and void student nets, respectively.
To obtain a warm-up-training-based student net, defenders create an encoder with randomly initialized parameters and then pre-train it with a small set of clean data by contrastive learning.
To evaluate the performance of different student nets comprehensively, we conduct experiments with multiple teacher nets together for each fixed student net.
For example, Table~\ref{tab:student_pertrained} presents the distillation results with warm-up-training-based encoders as the student nets.
The top half and bottom half of the table showcase backdoor mitigation performance against two state-of-the-art attacks, i.e., BadEncoder and BASSL.
The column ``Undefended'' denotes the raw input backdoored encoder. 
Columns ``T-FT'', ``T-FP'', ``T-ANP'', and ``T-MOTH'' represent that distillation takes FT, FP, ANP, and MOTH as teacher nets, respectively.
The `Undefended'' column displays the performance of downstream task models built on the raw backdoored encoder in terms of ACS and ASR. The subsequent columns (named ``T-FT'', ``T-FP'', ``T-ANP'', and ``T-MOTH'') exhibit the performance of downstream task models built on distilled encoders. These distilled encoders are initialized with warm-up-training-based encoders and trained by distilling knowledge from the teacher nets produced through FT, FP, ANP, and MOTH techniques, respectively. 
For instance, the GTSRB classifier built on BadEncoder-attacked CIFAR10 encoder (the first row) has 98.64\% ASR. 
Distillation employing warm-up-training-based student nets reduces the ASR to 4.72\% at best, albeit with an effectiveness loss of 93.92\%.
To find out which candidate is the best student net, we also compute the average performance of each candidate.
As shown in Figure~\ref{fig:performance_student_nets}, warm-up-training-based student nets achieve BS of 0.79, while raw backdoored student nets and void student nets' BS are 0.77 and 0.76, respectively.
In particular, raw poisoned student nets inherit the largest ASR of 19.48\%, while warm-up-training-based encoders have an ASR of 13.15\%.
To mitigate backdoor influence, void student nets have to suffer an ACC loss of 6.05\%, which warm-up-training-based student nets reduce 5.82\% accuracy.
Taking both into consideration, warm-up-training-based student nets reach BS of 0.79, while the second best is 0.77.
In that case,  obtaining student nets by warm-up training is the best choice of distillation during the evaluation.
Based on our analysis of the experiments, it's expected that warm-up-training-based student nets achieve the best performance.
When adopting raw backdoored encoders as student nets directly, the effectiveness is guaranteed, but the security is under great threat.
As declared in~\citep{2021-DFST}, the injected backdoor as a low-level feature (e.g. pixel perturbation) is more robust than these semantic high-level features.
In that case, the ASR of raw backdoored students doesn't meet the standard of security.
As for void student nets, though they achieve no backdoor inheritance, the effectiveness suffers a great loss.
Though void encoders avoid backdoor inheritance because of random initialization, benign knowledge is also thrown away in this process.
For instance, with void encoders as student nets and FT as teacher nets, the GTSRB classifier built on BASSL-attacked CIFAR10 encoder (the first row, bottom half of Table~\ref{tab:student_void_enc}) has 10.75\% ACC reduction, while the other two's ACC loss is less than 2\%.
To handle it, warm-up training improves simple random initialization by pre-training the void encoder on a small set of clean data.
In this experiment, we utilize contrastive loss help student nets learn image representations at first, which is proven effective by our experiments.

\emph{The impact on different distillation losses.}
Finally, with the optimal teacher and student networks determined, we evaluate the effect of different distillation losses on backdoor mitigation.
As established earlier, the fine-tuned (FT) teacher network achieves the highest performance among all teacher candidates (FT, FP, ANP, and MOTH), while the warm-up–trained student provides the best trade-off between effectiveness and security.
Under this configuration, we compare six representative distillation losses, including two feature-level losses (``F-FitNets'' ~\citep{romero2014fitnets}, ``F-CC''~\citep{peng2019correlation}), two attention-level losses (``AT-AFD''~\citep{2019-AFD}, ``AT-ATD''~\citep{2016ATD}), and two layer-level losses (``L-SP''~\citep{2019-SP}, ``L-KD''~\citep{hinton2015distilling}).
Feature-based losses transfer high-level representations from the teacher to the student, attention-based losses align spatial importance maps to guide the student’s focus, and layer-based losses match intermediate features for finer-grained supervision.
We conduct evaluations using the FT-based teacher and warm-up–trained student across one pre-training dataset and three downstream tasks, under both BadEncoder and BASSL threat models.

\begin{table*}[t]
    \centering
    \tabcolsep=2pt
    \renewcommand{\arraystretch}{1.1} 
  \caption{Effect of different losses on distillation.}
    \begin{tabular}{cccccccccccccccccc}
        \toprule

        \multirow{2.5}{*}{Attack} &
        \multirow{2.5}{*}{Pre-trian} &
        \multirow{2.5}{*}{Downstream} &
        \multicolumn{2}{c}{Undefended} &
         \multicolumn{2}{c}{F-FitNets} &
        \multicolumn{2}{c}{F-CC} &
        \multicolumn{2}{c}{AT-AFD} &
        \multicolumn{2}{c}{AT-ATD} &
        \multicolumn{2}{c}{L-SP} &
        \multicolumn{2}{c}{L-KD} &\\

        \cmidrule(lr){4-5} \cmidrule(lr){6-7} \cmidrule(lr){8-9} \cmidrule(lr){10-11} \cmidrule(lr){12-13} \cmidrule(lr){14-15} \cmidrule(lr){16-17}
        & & & ACC & ASR & ACC & ASR & ACC & ASR & ACC & ASR & ACC & ASR & ACC & ASR & ACC & ASR\\
            
        \cmidrule(lr){1-17}
        
        \multirow{4}{*}{\rotatebox{90}{BadEncoder}} & \multirow{3}{*}{CIFAR10} & GTSRB  & 82.27 & 98.64 & \textbf{77.53} & 4.59 & 74.47 & 4.71 & 42.62 & 3.99 & 75.36 & 5.06  & 70.48 & \textbf{3.91} & 73.78 & 4.81\\
        
        & & SVHN  & 69.32 & 99.14 & 58.01 & 24.38  & 51.96 & 45.95 & \textbf{65.83} & 37.62  & 63.92 & 28.53  & 54.86 &  \textbf{22.08} & 58.06 & 44.22\\
        
        & & STL10  & 76.18 & 99.17 & \textbf{69.48} & 18.88  & 67.10 & 18.28 & 58.18 & \textbf{7.06}  & 67.16 & 12.23 & 62.66 &  10.57 & 61.58 & 11.32  \\

        \cmidrule(lr){2-17}

        & \multicolumn{2}{c}{Average}  & 75.92 & 99.17 & 68.33 & 15.94  & 64.51 & 22.98 & 55.54 & 16.22 & \textbf{68.81}  & 15.27 & 62.66 &  \textbf{12.18} & 64.47 & 20.11\\ 
        
        \cmidrule(lr){1-17}
        
        \multirow{4}{*}{\rotatebox{90}{BASSL}} & \multirow{3}{*}{CIFAR10} &  GTSRB  & 78.96 & 66.12 & \textbf{78.48} & 9.05 & 74.10 & 5.84 & 12.88 & \textbf{0.20} & 77.49 & 2.09 & 71.22 & 33.58  & 72.18 & 5.14\\
        
        & & SVHN  & 66.54 & 81.97 & 63.34 & 15.31 & 54.22 & 11.19 & 58.53 & 11.8 & \textbf{65.86} & 13.12  & 55.03 &  \textbf{3.22} & 58.32 & 25.32\\
         
        & & STL10  & 72.99 & 37.94 & \textbf{69.95} & 11.41 & 66.80 & \textbf{9.12} & 57.29 &  6.39 & 67.27 & 11.41 & 62.30 &  12.38 & 62.26 & 12.76\\

        \cmidrule(lr){2-17}

        & \multicolumn{2}{c}{Average}   & 72.83 & 62.01 & \textbf{70.58} & 11.92 & 65.04 & 8.71 & 42.90 & 6.13 & 70.20  & \textbf{8.87} & 62.85 &  16.39 & 64.25 & 14.40\\ 
        \bottomrule
    \end{tabular}
\label{tab:distillation_loss}
\end{table*}

Table~\ref{tab:distillation_loss} summarizes the performance of distillation under six loss functions using the same fine-tuned (FT) teacher and warm-up–trained student networks.
Among all loss candidates, the attention-based AT-ATD achieves the best overall performance, reducing ASR from 80.59\% to 12.07\% while incurring only a 4.88\% drop in accuracy.
In contrast, the second-best method retains an ASR of 13.93\% and loses 4.92\% accuracy.
Aggregating by category, attention-based losses reduce ASR to 11.62\% on average, outperforming feature-level (14.88\%) and layer-level (15.77\%) alternatives.
We attribute this advantage to the spatial alignment introduced by attention, which enables the student network to better capture the teacher’s benign focus regions while discarding poisoned feature dependencies.
However, not all attention losses behave equally—AFD performs worse due to its reliance on a small subset of high-activation neurons, which may include backdoored units and distort benign feature alignment.

\smallskip
\noindent
\textbf{RQ3: How robust and generalizable is distillation against variations in trigger patterns, model architectures, and pre-training algorithms?}

We further examine the robustness and generalization of distillation across diverse settings. Specifically, we evaluate its performance under varying trigger sizes and model architectures.
The trigger size reflects the attack strength, i.e., larger triggers generally make backdoors more difficult to remove, while different architectures simulate heterogeneous real-world deployment scenarios. Demonstrating consistent effectiveness across these conditions provides strong evidence that distillation is a robust and generalizable defense against backdoor attacks.

\emph{The analysis of trigger size.}
\label{subsubsec:RQ4_1}
Attackers try to implant a pre-defined trigger (e.g., square patch) in pre-trained encoders, aiming to manipulate specific downstream tasks.
The size of the injected trigger corresponds to the pixel block dimensions.
Trigger size affects the attack's effectiveness and stealthiness.
The larger the injected trigger is, the easier for a successful attack.
The smaller the injected trigger is, the more stealthy an attack, but easier for defenders to reduce the ASR.
It's mitigating backdoors with a large trigger,  thereby substantiating the robustness of the distillation process. 

\begin{figure}[t]
\centering
\begin{subfigure}{0.49\columnwidth}
    \includegraphics[width=\linewidth]{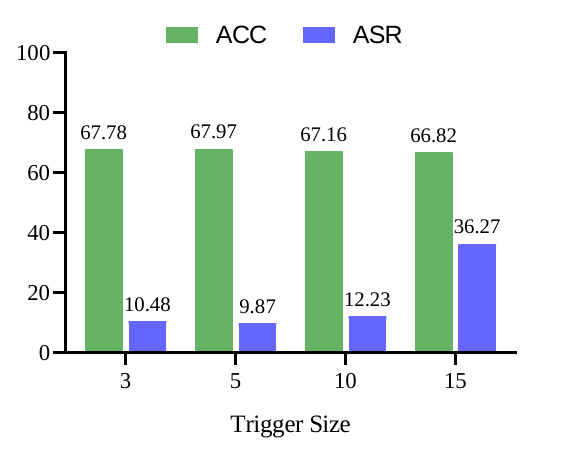}
    \caption{STL10}
    \label{fig:trigger_size_STL10}
\end{subfigure}
\hspace{-6mm}
\begin{subfigure}{0.49\columnwidth}
    \includegraphics[width=\linewidth]{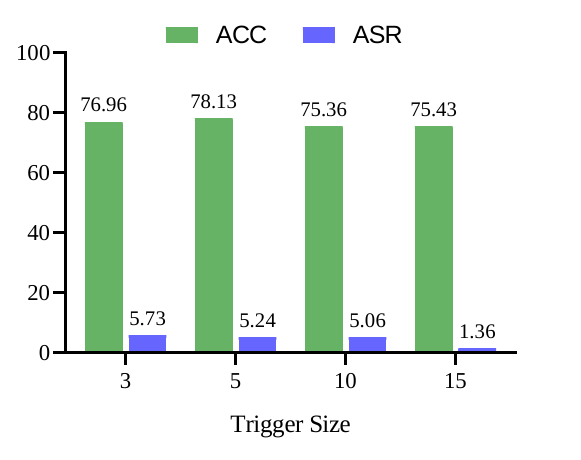}
    \caption{GTSRB}
    \label{fig:trigger_size_GTSRB}
\end{subfigure}
\caption{Impact of trigger size.}
\label{fig:influence_trigger_size}
\end{figure}

\begin{figure}[t]
\centering
    {
        \includegraphics[width=0.3\linewidth]{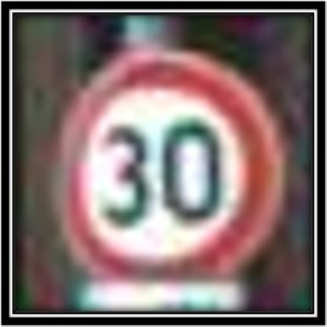}
    }
    {
        \includegraphics[width=0.3\linewidth]{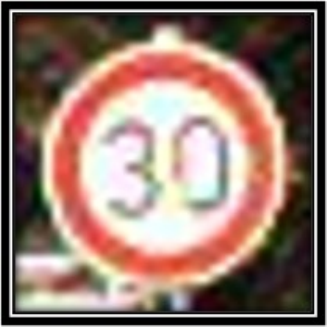}
    }
    {
        \includegraphics[width=0.3\linewidth]{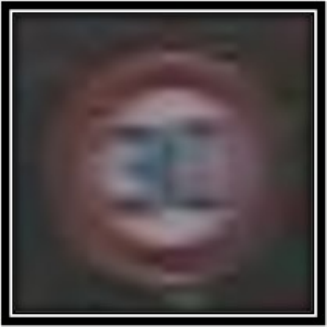}
    }
     
    \caption{Representative traffic sign images from the GTSRB dataset, illustrating variations in illumination, resolution, and viewing angle.}
    \label{fig:samples_from_GTSRB}
\end{figure}

We conduct experiments on four different sizes, including 3 × 3, 5 × 5, 10 × 10, and 15 x 15.
Figure~\ref{fig:influence_trigger_size} reports the experimental results, where the green bar and blue bar indicate the ACC and ASR after distillation, respectively.
We attack pre-trained encoders with BadEncoder~\citep{2022-BadEncoder} on two different downstream tasks, GTSRB and STL10.
Both encoders are pre-trained on CIFAR10.
Figure~\ref{fig:trigger_size_STL10} represents the experiments with STL10 as the downstream task.
It's reported that distillation removes backdoors totally even with a trigger size of 10$\times$10.
When the trigger size comes to 15$\times$15, the ASR after a distillation still remains 36.27\%.
And it's illustrated that distillation performs well even if the trigger sizes vary in GTSRB.
Based on our analysis, the results in STL10 are in line with our expectations.
With a trigger size of 15$\times$15, it occupies a quarter of a given image.
The backdoor of such a trigger is too strong to be removed. 
The results of GTSRB can attributed to its simplicity.
GTSRB contains 43 classes and it's obtained by data argumentation (Figure~\ref{fig:samples_from_GTSRB}). 
In that case, it's easier for distillation to separate benign knowledge and obtain a cleansed encoder.

\emph{The analysis of model architecture.}
This experiment aims to assess the effectiveness of distillation for backdoor mitigation across varying model architectures.
Performance metrics, ACC, and ASR, will be evaluated to elucidate the impact of distillation on different ResNet architectures, providing valuable insights into the generalization capabilities and optimization potential across the model spectrum.

\begin{table}[t]
\centering
\tabcolsep=5pt
\renewcommand{\arraystretch}{1.1} 
\caption{Results on different pre-training architectures.}

\begin{tabular}{lcccc}
\toprule
\multirow{2}{*}{Architecture} & \multicolumn{2}{c}{Undefended} & \multicolumn{2}{c}{Distillation} \\
\cmidrule(lr){2-3} \cmidrule(lr){4-5} 
 & ACC & ASR & ACC & ASR \\
\midrule
ResNet18 & 77.78 &99.76 &  71.92& 24.37\\ 
ResNet34 & 77.41 &99.86 & 70.36 &10.05  \\
ResNet50 & 77.71 &99.99 & 72.41 & 12.27 \\
\bottomrule
\end{tabular}

\label{tab:generalization_archs}
\end{table}

Table~\ref{tab:generalization_archs} shows distillation performance across varied model architectures, including ResNet18(RN18), ResNet34(RN34) ,and ResNet50(RN50).
With undefended backdoored encoders, RN18 achieves an accuracy of 77.78\% with an ASR of 99.76\%, while RN34's specific metrics are 77.41\% and 99.86\%, respectively.
RN50 attains an accuracy of 77.71\% with an ASR of 99.99\%.
After distillation, the cleansed encoders mitigate ASRs to 24.37\%, 10.05\%, and 12.27\% under RN18, RN34, and RN50, respectively.
The ACC reduction is controlled within 10\% on average.
It's observed that distillation has stable backdoor removal performance on multiple architectures.
Distillation also shows its performance on deeper and larger models, RN34 and RN50, which is better than RN18.
That's because our distillation is deployed on the level of layers.
Deeper and larger models mean more layers, leading to higher distillation strength. 
For example, the distillation loss consists of four parts for a four-layer model.
Given an eight-layer model, the distillation loss contains eight parts, which makes distillation more effective in removing backdoors.

\emph{The analysis of pre-training algorithm.}
Pre-training algorithms are used to pre-train encoders, which adversaries can then exploit.
They attack encoders pre-trained with multiple pre-training algorithms, which are more likely to be used in downstream tasks.
We therefore analyze distillation performance when it's applied to encoders pre-trained by different algorithms.
We conduct an experiment with encoders pre-trained by SimCLR~\citep{chen2020simple}, BYOL~\citep{grill2020bootstrap}, and MoCo-v2~\citep{chen2020mocov2}.

\begin{table}[tbp]
\centering
\tabcolsep=5pt
\renewcommand{\arraystretch}{1.1} 
\caption{Results on different pre-training algorithms.}

\begin{tabular}{lcccc}
\toprule
\multirow{2}{*}{Algorithm} & \multicolumn{2}{c}{Undefended} & \multicolumn{2}{c}{Distillation} \\
\cmidrule(lr){2-3} \cmidrule(lr){4-5} 
 & ACC & ASR & ACC & ASR \\
\midrule
SimCLR & 77.78 & 99.76 & 71.92 & 24.37 \\
MoCo-v2 & 72.85 & 99.98 & 70.70 & 26.83 \\
BYOL & 75.63 & 99.98 & 73.42 &  22.95 \\
\bottomrule
\end{tabular}

\label{tab:generalization_methods}
\end{table}

Table~\ref{tab:generalization_methods} reports the experimental results.
Column ``Algorithm'' means different pre-training methods.
Columns ``Undefended'' and ``Distillation'' denote backdoored and distilled encoders, respectively.
Observe that distillation can effectively mitigate backdoors in a variety of pre-training algorithms, delineating its generalizability.
The performance of distillation with different pre-training algorithms demonstrates that (1) distillation can mitigate backdoors even with different pre-training algorithms, (2) distillation performance varies when pre-training algorithm changes, resulting in 7.53\%, 2.95\%, 2.92\% decreases in ACC and 75.57\%, 73.16\%, 77.05\% decreases in ASR with SimCLR, MoCo-v2, and BYOL.

\smallskip
\noindent
\textbf{RQ4: How effective is distillation against advanced backdoor attacks?}
\begin{table}[tbp]
\centering
\tabcolsep=5pt
\renewcommand{\arraystretch}{1.1} 
\caption{Results of distillation on advanced attacks.}

\begin{tabular}{lcccccc}
\toprule
\multicolumn{1}{l}{\multirow{2}{*}{Attack}} & \multicolumn{2}{c}{Undefended} & \multicolumn{2}{c}{Teacher} & \multicolumn{2}{c}{Distillation} \\ 
\cmidrule(lr){2-3} \cmidrule(lr){4-5} \cmidrule(lr){6-7} 
\multicolumn{1}{l}{} & ACC & ASR & ACC & ASR & ACC & ASR \\
\cmidrule(lr){1-7} 
CTRL  & 71.67 & 44.05 & 67.13 & 11.18 & 68.96 &  9.68 \\
BLTO  & 76.57 & 52.45 & 71.41 & 53.76 & 69.88 & 40.95 \\
DRUPE & 72.68 & 76.68 & 69.12 & 14.35 & 68.92 & 11.37 \\ 
\bottomrule
\end{tabular}

\label{tab:advance_attack}
\end{table}

To go beyond BadEncoder and BASSL, we further evaluate distillation under three recently proposed advanced SSL backdoor attacks: CTRL~\citep{2023-CTRL}, BLTO~\cite{2024-BLTO}, and DRUPE~\cite{2024-DRUPE}. 
CTRL~\cite{2023-CTRL} leverages contrastive learning to inject backdoors into SSL models, using only a small amount of poisoned data to achieve strong and stable attack effects, thereby revealing the high vulnerability of SSL to backdoor threats.
BLTO~\citep{2024-BLTO} further enhances stealth and robustness by jointly optimizing the trigger and the contrastive feature space, making the backdoor more adaptive and difficult to remove.
DRUPE~\citep{2024-DRUPE} focuses on preserving the feature distribution of clean samples, so that poisoned data remain statistically indistinguishable, which significantly increases the challenge for defenses.

Experimental results in \Cref{tab:advance_attack} show that although these three attacks are much stronger and more stealthy than earlier ones, distillation still effectively mitigates their impact. For example, under CTRL, BLTO, and DRUPE, distillation reduces the ASR from 44.05\%, 52.45\%, and 76.68\% to 9.68\%, 40.95\%, and 11.37\%, respectively, while maintaining comparable accuracy. 
Note that the ASR remains relatively high under BLTO.
This can be attributed to its bi-level optimization mechanism, which jointly optimizes the trigger and representation spaces, thereby entangling backdoor features with benign semantics and making them harder for distillation to disentangle and remove.
This observation suggests a promising future direction: developing distillation frameworks that explicitly separate or regularize feature subspaces, enabling more effective purification of semantically entangled backdoors.
Overall, the above results demonstrate that distillation retains strong backdoor removal capability even against advanced self-supervised learning backdoor attacks, highlighting its great potential as a general and robust defense mechanism.

\section{Conclusions}
\label{sec:conclusion}
In this paper, we perform the first exploration into the effectiveness of distillation in mitigating backdoor attacks for SSL. Specifically, we undertake rigorous empirical experiments to answer three research questions, systematically investigating four facets of distillation. We empirically demonstrate that distillation can, to a large extent, mitigate backdoor influence in pre-trained encoders in SSL. We find that the combination of fine-tuned teacher nets, warm-up training-based student nets, and attention-based distillation loss achieves the best performance in the framework. We empirically validate some potential improvements to promote distillation performance. It is found that more clean data helps distillation to anchor benign knowledge in teacher nets. We find that distillation maintains robustness and generalization under different trigger sizes and model architectures, highlighting further the exploration of synergies between distillation and other defense mechanisms to create more robust and resilient machine learning systems.

\section{Data availability}

All source code and experimental data used in this study are publicly available at~\url{https://github.com/wssun/SSLBackdoorMitigation}.

\section*{Acknowledgments} 
This research is supported partially by the National Natural Science Foundation of China (62141215, 61932012, 62372228), and the National Research Foundation, Singapore, and the Cyber Security Agency under its National Cybersecurity R\&D Programme (NCRP25-P04-TAICeN). Any opinions, findings and conclusions or recommendations expressed in this material are those of the author(s) and do not reflect the views of National Research Foundation, Singapore and Cyber Security Agency of Singapore.

\bibliographystyle{spbasic} %
\bibliography{main}%

\end{document}